\documentclass[
anonymous
]{ceurart}

\usepackage{listings}
\usepackage{hyperref}
\begin{document}

\copyrightyear{2023}
\copyrightclause{Copyright for this paper by its authors. Use permitted under Creative Commons License Attribution 4.0 International (CC BY 4.0).}

\conference{ISWC 2023 Posters and Demos: 22nd International Semantic Web Conference, November  6–10, 2023, Athens, Greece}

\title{DBLPLink: An Entity Linker for the DBLP Scholarly Knowledge Graph}


\author[1]{Debayan Banerjee}[%
orcid=,
email=debayan.banerjee@uni-hamburg.de,
url=,
]

\author[2]{Arefa}[%
orcid=,
email=arefa.muzaffar@gmail.com,
url=,
]

\author[1]{Ricardo Usbeck}[%
orcid=,
email=ricardo.usbeck@uni-hamburg.de,
url=,
]
\author[1]{Chris Biemann}[%
orcid=,
email=chris.biemann@uni-hamburg.de,
url=,
]

\address[1]{Universität Hamburg, Hamburg, Germany}
\address[2]{Jamia Milia Islamia, New Delhi, India}


\begin{abstract}
  In this work, we present a web application named DBLPLink, which performs entity linking over the DBLP scholarly knowledge graph. DBLPLink uses text-to-text pre-trained language models, such as T5, to produce entity label spans from an input text question. Entity candidates are fetched from a database based on the labels, and an entity re-ranker sorts them based on entity embeddings, such as TransE, DistMult and ComplEx. The results are displayed so that users may compare and contrast the results between T5-small, T5-base and the different KG embeddings used. The demo can be accessed at \href{https://ltdemos.informatik.uni-hamburg.de/dblplink/}{https://ltdemos.informatik.uni-hamburg.de/dblplink/}. Code and data shall be made available at  \href{https://github.com/uhh-lt/dblplink}{https://github.com/uhh-lt/dblplink}.
\end{abstract}


\maketitle
 \vspace{-4mm}
\section{Introduction and Related Work}
Entity Linking (EL) is a natural language processing (NLP) task that involves associating named entities mentioned in text to their corresponding unique identifiers in a knowledge graph (KG). For example, in the question: \textit{Who is the president of USA?}, the named entity span of \textit{USA} has to be linked to the unique identifier \texttt{Q30}\footnote{\url{https://www.wikidata.org/wiki/Q30}} in the Wikidata KG~\cite{vrandevcic2014wikidata}. Several entity linkers exist~\cite{Sevgili2022} over general purpose KGs such as Wikidata, and  more specialized KGs, such as bio-medical~\cite{french2023overview} or financial KGs~\cite{elhammadi-etal-2020-high}, however, to the best of our knowledge, no working entity linker exists for scholarly KGs. 

A scholarly KG is a special sub-class of KGs, which contains bibliographic information about research publications, authors, institutions etc. Some well-known scholarly KGs are the OpenAlex\footnote{\url{http://openalex.org/}}, ORKG\footnote{\url{https://orkg.org/}} and DBLP\footnote{\url{https://dblp.org/}}. In this work, we focus on the DBLP KG, which caters specifically to computer science, and as a result, is smaller in size than other scholarly KGs. DBLP, which used to stand for Data Bases and Logic Programming\footnote{\url{https://en.wikipedia.org/wiki/DBLP}}, was created in 1993 by Michael Ley at the University of Trier, Germany~\cite{leyDBLPComputerScience2002}. At the time of its release\footnote{\url{https://blog.dblp.org/2022/03/02/dblp-in-rdf/}}, the RDF dump consisted of 2,941,316 person entities, 6,010,605 publication entities, and 252,573,199 RDF triples.

DBLPLink can handle simple and complex questions pertaining to authorship, venues, institutions and other information available in the DBLP KG. 

\section{Web Interface}

\begin{figure*}[ht]
\vspace{-4mm}
\centering
    \includegraphics[width=1.0\linewidth]{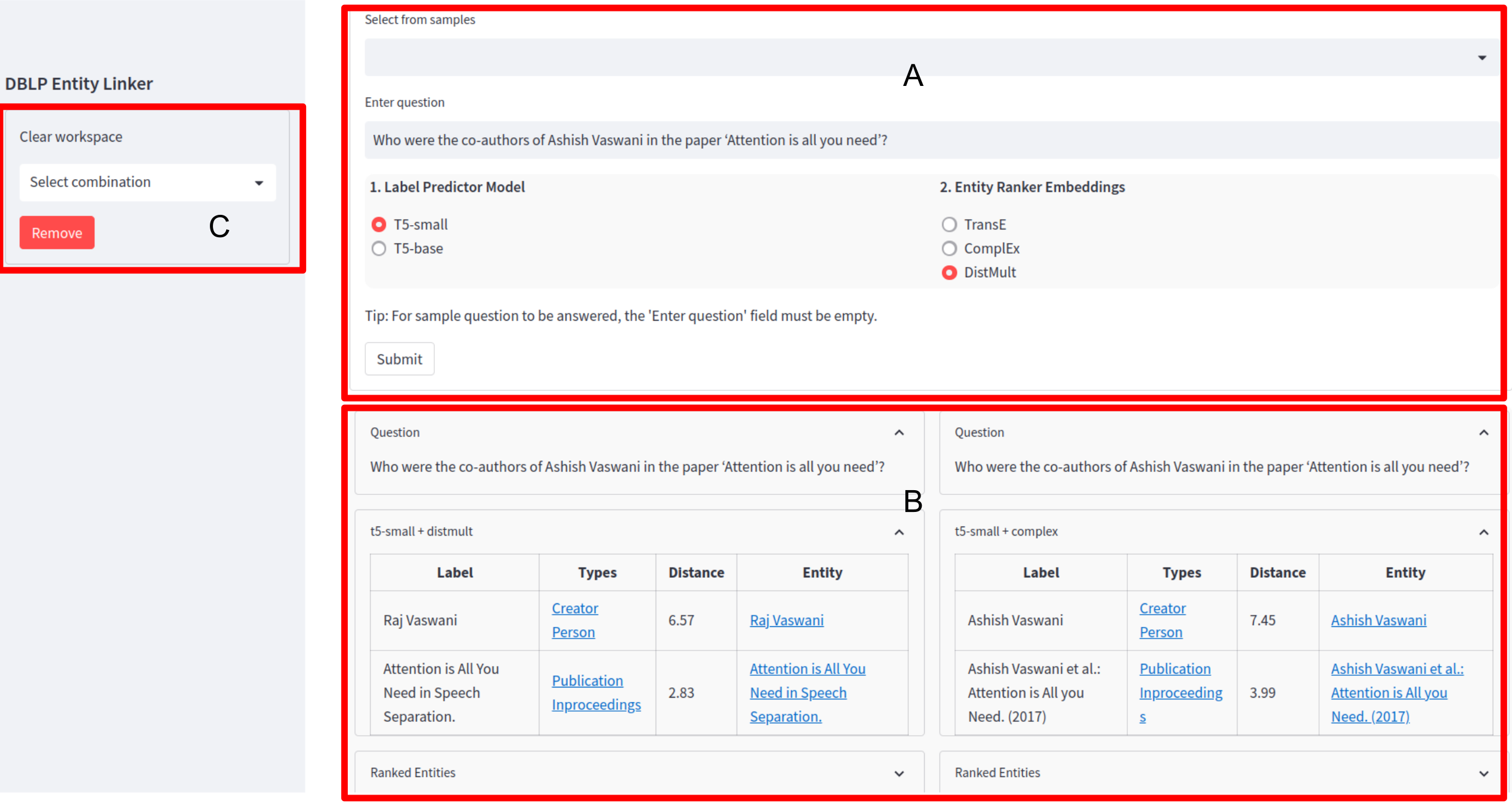}
\caption{User interface of DBLPLink. The question reads: "Who were the co-authors of Ashish Vaswani in the paper 'Attention is all you need'?"}
\label{ui1}
\vspace{-4mm}
\end{figure*}

As shown in Figure \ref{ui1}, the UI consists of three main parts. In \textbf{Section A}, the user can either type a question as input or select a question from the drop-down menu. Further, the user can select which model to use for label span detection, and which embeddings to use for re-ranking of entities. In \textbf{Section B}, the results of DBLPLink are displayed. First, the top-ranked entity for each detected span is displayed, with a corresponding label and type from the DBLP KG. A hyperlink to the entity, which points to the original DBLP entity web page is also shown. Additionally, a \texttt{distance} metric is shown which denotes how close a match this entity is to the input question. A lower distance means a better match. Towards the bottom of the UI, we can briefly see collapsible boxes called "Ranked Entities", which further display the top 10 ranked entities for each of the detected label spans. Lastly, in \textbf{Section C}, the user has an option to remove certain combinations of results from the screen, if the UI becomes too cluttered. Our expectation is that the user shall try multiple combinations of T5 and entity embeddings to compare and contrast the results, which may need occasional cleanup from the UI.

\begin{figure*}[ht]
\centering
    \includegraphics[width=1.0\linewidth]{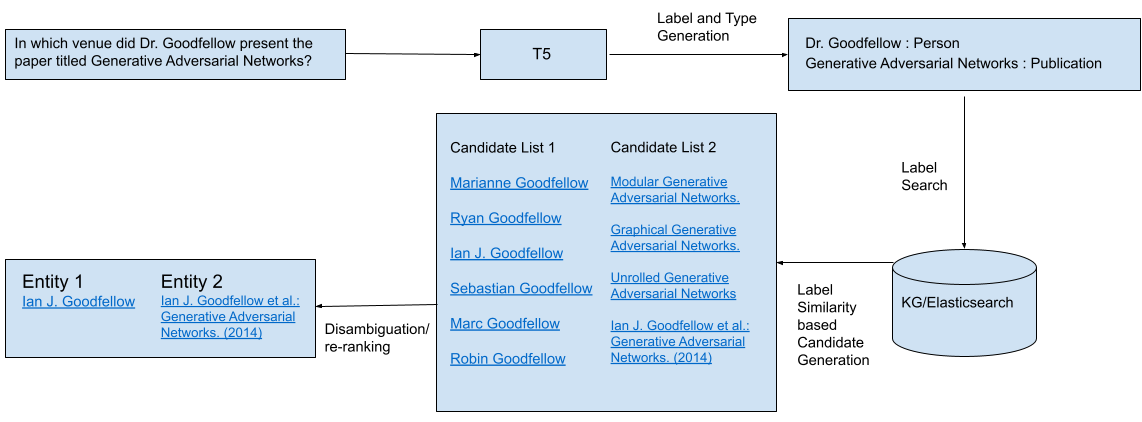}
\caption{Architecture of DBLPLink.}
\label{ui2}
\end{figure*}



 \vspace{-4mm}
\section{Architecture}
 \vspace{-2mm}
\subsection{Label and Type Generation}
\vspace{-2mm}

As seen in Figure~\ref{ui2}, the first step is to produce salient labels and types from the given input question. For this purpose, we use the DBLP-QuAD~\cite{dblpquad} dataset to fine-tune T5-small and T5-base~\cite{t5} models, on the task of producing entity labels and types from the input question. 

\vspace{-4mm}
\subsection{Candidate Generation}
\vspace{-2mm}

With the entity labels and types produced in the previous step, a free-text-search is performed on an Elasticsearch\footnote{\url{https://www.elastic.co/}} instance, which contains entity URLs with their corresponding labels. The results are further filtered by the types. This gives us a list of candidate entities. In normal operation of the demo application, we present the top-ranked candidate as the final linked entity. We only proceed to the disambiguation stage if the top entity candidate has a label, that is the same as another entity in the candidate list. 

\vspace{-4mm}
\subsection{Disambiguation}
\vspace{-2mm}

In case two entities in the candidate list share the same label, we proceed with disambiguation, which requires a further re-ranking of the candidate list. For this, we follow a common approach of using Siamese neural networks~\cite{DBLP:conf/nips/BromleyGLSS93} for learning text similarity between text pairs \cite{ranasinghe-etal-2019-semantic}. We embed the input question and the candidate entities in a common embedding space. For this purpose, we create a 969-dimensional embedding, where for a given question, we use the first 768 dimensions for the BERT embedding. We fill the remaining 201 dimensions with zeros. For the entity candidates, we fill the first 768 dimensions with the BERT embedding of the entity label, while the next 200 dimensions are reserved for the entity embeddings. We use three different kinds of embeddings in our experiments, namely TransE~\cite{10.5555/2999792.2999923}, ComplEx~\cite{pmlr-v48-trouillon16}, and DistMult~\cite{yang2015embedding}. 
For the remaining 969th dimension, we store the degree of string similarity match between the entity label and the input question. For training, pairs of positive and negative samples are used with a triplet ranking loss function and L2 distance metric.

During inference, a question and an entity candidate are vectorised and passed through the trained Siamese network. The cosine distance between the two resulting embeddings is computed, and the pair with the lowest distance is considered the most suitable match.



\begin{table*}[]
\vspace{-4mm}
  \centering
  
  \begin{tabular}{|c|c|c|c|c|c|c|c|}
  \hline
   & \multicolumn{1}{c|}{}&\multicolumn{3}{c|}
   {conditional-disambiguation}&\multicolumn{3}{c|}
   {hard-disambiguation} \\
    \hline
      &Label Sorting&TransE&ComplEx&DistMult&TransE&ComplEx&DistMult\\
      \hline
    
    \hline
     T5-small          & 0.698 & 0.700 & 0.692& 0.699&0.511 & 0.482 & 0.537\\
     T5-base           & 0.698  & \textbf{0.701} & 0.692 & \textbf{0.701} & 0.521& 0.484& 0.547 \\

    \hline
  \end{tabular}
  
  \caption{F1-scores for the entity linking task across different combinations of span detector and entity re-ranker}
  \label{table1}
  \vspace{-4mm}
\end{table*}

 \vspace{-4mm}
\section{Evaluation}
 \vspace{-4mm}
We evaluate our entity linker on the 2.000 questions of the test split of the DBLP-QuAD dataset and measure the F1-score. In Table \ref{table1}, under the heading `Label Sorting`, we consider the top-ranked candidate after the label sorting phase as the linked entity. We perform no further disambiguation. Under the `conditional-disambiguation` setting, we perform disambiguation only if two entities in the candidate list share the same label.  Under the `hard-disambiguation` setting, re-ranking based on Siamese network cosine distances is always run after the candidate generation phase, essentially ignoring the label sorting order.

We see that hard-disambiguation lags behind significantly in performance when compared to plain label sorting, which points to the learning that for DBLP KG, degree of string match of an author or a publication is more important than the KG embeddings. Based on this finding, we allow the web application to run in `conditional-disambiguation` mode for better performance. In the case of conditional disambiguation, performance is marginally better when using TransE and DistMult when compared to label sorting, because not many cases of ambiguous labels exist in the DBLP-QuAD test set. However, it is evident from the hard disambiguation case, that DistMult performs the best on a pure disambiguation task. This may be explained by the inherent suitability of DistMult for 1-to-N relationships, which is close to the nature of the DBLP KG model, where one author may have several papers. On the contrary, TransE expects 1-to-1 relationships, while ComplEx works better for symmetric relationships. Another interesting outcome of the experiments is that the difference in parameter sizes of T5-small and T5-base does not produce any difference in performance. This may be explained by the fact that in the span label production task, much of the focus is on copying the right part of the input to the output. Since the learned knowledge of the model weights from the pre-training task is not being exploited, the larger size of T5-base does not seem to matter.

 \vspace{-4mm}
\section{Conclusion}
 \vspace{-2mm}
In this work, we presented DBLPLink, which is a web-based demonstration of an entity linker over the DBLP scholarly KG. In the future, we would like to add further interactivity to the UI where users can provide feedback on quality of the results. Additionally, a conversational interface for question answering would be desirable for question answering tasks, and we would like to build it in a future version.



\vspace{-3mm}
\section{Acknowledgements}
\vspace{-3mm}

This research is performed as a part of the ARDIAS project, funded by the “Idea and Venture Fund“ research grant by Universität Hamburg, which is part of the Excellence Strategy of the Federal and State Governments. This work has additionally received funding through the German Research Foundation (DFG) project NFDI4DS (no.~460234259).
\bibliography{sample-ceur}

\end{document}